# Adaptive Real-Time Removal of Impulse Noise in Medical Images

Zohreh HosseinKhani[1], Mohsen Hajabdollahi[1], Nader Karimi[1], S.M. Reza Soroushmehr[2,3], Shahram Shirani[4], Kayvan Najarian[2,3], Shadrokh Samavi[1,4]

[1]Department of Electrical and Computer Engineering, Isfahan University of Technology, Isfahan 84156-83111, Iran
[2]Michigan Center for Integrative Research in Critical Care, University of Michigan, Ann Arbor, MI 48109 U.S.A
[3]Department of Emergency Medicine, University of Michigan, Ann Arbor, MI 48109 U.S.A
[4]Department of Electrical and Computer Engineering, McMaster University, Hamilton, ON L8S 4L8, Canada

**Abstract:** Noise is an important factor that degrades the quality of medical images. Impulse noise is a common noise, which is caused by malfunctioning of sensor elements or errors in the transmission of images. In medical images due to presence of white foreground and black background, many pixels have intensities similar to impulse noise and distinction between noisy and regular pixels is difficult. In software techniques, the accuracy of the impulse noise removal is more important than the algorithm's complexity. But in development of hardware techniques having a low complexity algorithm with an acceptable accuracy is essential. In this paper a low complexity de-noising method is proposed that removes the noise by local analysis of the image blocks. In this way, noisy pixels are distinguished from non-noisy pixels. All steps are designed to have low hardware complexity. Simulation results show that in the case of magnetic resonance images, the proposed method removes impulse noise with an acceptable accuracy.

I. INTRODUCTION

Noise is introduced in medical images either during the image capture process, during compression and image processing routines, or during the transmission of the images. Introduction of noise degrades the quality of the images. One of the most common noises is salt & pepper noise, which is a member of the broader category of impulse noises. Salt & pepper consist of two constant values which are distributed randomly throughout the image. In order to remove salt & pepper noise many studies have been performed which consist of two stages. The two main stages include detection of a noisy pixel and replacement of the noisy pixel with a proper value. We will review some of the existing noise removal methods. We initially look at general purpose methods and then we will review some methods that are specifically designed for medical images.

Some methods are simple and suitable for hardware implementation while some others are complex. Proposed methods in [1-6] can be considered as complex algorithms, in terms of hardware implementation. In [1] in order to detect noisy pixels, image histogram and fuzzy method are used. Then, for the restoration stage, a median filter is applied around the noisy pixel. In [2], by using adaptive fuzzy method, noisy pixels are detected and restored with weighted mean filter. In [3], an uncertainty based detector finds the noisy pixel. Then a weighted fuzzy filter is applied and removes the noise. In [4] a de-noising method based on a second generation wavelet as well as adaptive median filtering is used for noise removal. In [5] a de-noising method is proposed which noisy pixels are detected by comparing them with max and min values of the gray scale image. Then, a spline interpolation function is applied on the non-noisy pixels. In [6] an evolutionary algorithm and an improved median operation are used for the detection and restoration steps respectively.

On the other hand, the proposed methods in [7-10] can be considered as suitable methods for hardware implementation. In [7] for impulse noise removal, four edge directions are considered in $5 \times 5$ windows. According to the difference between pixels in each direction, noisy pixels are detected. With median operation on non-noisy pixels, restoration is performed. In [8] a $3 \times 3$ block around each pixel is considered and it is sorted in all directions. Then maximum, minimum and median values of the $3 \times 3$ block are computed for noisy pixel detection. Restoration is performed with median operation on the non-noisy pixels. In [9] for detection of the noisy regions, similarity between a pixel and noisy pixel is computed using the Laplacian operator with a specific threshold. Based on the values of neighboring pixels, noisy pixels are detected and vector median filtering is used for restoration. In [10] a decision based algorithm is proposed for enhancing images and videos which are corrupted by high density salt and peppers noise. In a neighborhood window, a pixel with the value of 0 or 255 is considered as a noisy pixel. A noisy pixel and its four main neighboring pixels are considered and the noisy pixel is restored by median or mean filtering.

Magnetic resonance (MR) images are affected by different noise sources, such machine generated artifacts, patient motion, signal processing noise, etc. [11]. Noise, in MR images, could occur even if the scanner has high resolution. Signal to noise ratio and visual quality are affected by the added noise. MR images contain different types of noise from various sources including abrupt changing, high physiological processing, eddy current, rigid body motion, non-rigid body motion and other sources [12]. It is necessary to identify and detect these types of noises to improve human body diagnostic method. In [13] a two-step algorithm for removal of Rician noise in MR images is proposed. Four types of filters are proposed and for all of them non-local linear minimum mean square error (LMMSE) estimation is used. In [14] Zernike moments are used based on non-local mean for denoising of Rician noise in MRI. Similar patches are found using non-local mean (NLM). Using Zernike moments, structure and edges of image are preserved. Then setting up a similarity metric, and operating same as NLM makes denoising method suitable for MR images.

Gaussian and impulse noises, which are created by malfunctions of electrical circuits and imaging devices, are the dominant types of distortions in medical imaging [15]. Presence of noise in MR images not only affects the quality of images but it also ruins the results of image enhancement techniques [11]. In [16] a fuzzy median filtering for the removal of impulse noises in MR images is proposed. Although the preservation of details in MR images is of major concern, but high computational complexity of [16] makes it unsuitable for hardware implementation. In [17] a neuro-fuzzy approach, which is an enhanced version of [16], is proposed. They use adaptive median filtering and many fuzzy rules are used to remove impulse noises.

The need for real-time implementation of some image processing applications makes hardware techniques more desirable and more applicable. For example in [18], the maximum and minimum values in a 3×3 window are calculated. Then edge directions are considered and noisy pixels are restored in the correct edge direction. Consideration of different directions, averaging, and differencing operations in all directions, make this algorithm relatively complex. In [19] a noisy pixel detection method, with variable window-size and a weighted filtering method, is proposed. In [20], for detection of random-value noisy pixels, a decision-tree is used and edge direction is similar to [18]. For medical image processing, hardware platforms, such as FPGAs and GPUs, are also considered [21].

In this paper, we are proposing a low complexity method for removal of impulse noise in medical images. The proposed method is suitable for hardware implementation demanded in many medical instruments. For detection of noisy pixels, which have values of either 0 or 255, similarity between its neighbors is considered. A pixel, which is not similar to its neighbors, is labeled as a noisy pixel. In the second stage for the reconstruction of noisy pixels, median filter is applied only on non-noisy pixels. For each stage an efficient hardware structure is proposed which makes the proposed method suitable for hardware implementation in medical devices. The major contributions of our proposed method are as follows:

- Distinction of noisy pixels from normal pixels with 255 (maximum white) or 0 intensities.
- Efficient hardware structure and its implementation on FPGA.

The rest of this paper is organized as follows. In Section 2, the proposed method for removal of impulse noise, including a software algorithm and its hardware architecture, are explained. Section 3 is dedicated to simulation results, and after that, in Section 4 concluding remarks are presented.

II. PROPOSED METHOD

In all real-time applications and especially in the case of noise removal, it is necessary to apply an efficient and accurate algorithm. The noise removal procedure can be considered as a preprocessing stage for many image processing applications. Complex software methods, such as neural networks and learning techniques, have been simplified for de-noising applications, where they have high accuracy and low complexity. Our proposed method, which is explained in the followings, has low complexity and good accuracy:

## A. General structure of the algorithm

The proposed method consists of a stage for noisy pixel detection, and another stage for replacement of the noisy pixel with a suitable value. The dataflow of the proposed method is displayed in the block diagram of Fig. 1. Also a graphical example of the proposed method is illustrated in Fig. 2. The proposed method consists of the following steps:

### 1) Pixel Labeling

In the first step of our proposed algorithm, pixels are labeled. Images are assumed to be gray-scale with pixel values between 0 and 255. An example of noisy MR image is shown in Fig. 2(a). A small region of the image is zoomed out and its pixel values are shown in Fig. 2(b). Label "0" is for pixels with zero value, "1" is for pixels with intensity of 255, and label "2" is used for pixels with any other intensity values. Results of the labeling process are illustrated in Fig. 2(c).

### 2) Noise-Free Pixel Detection

In the second stage of the proposed algorithm, noise-free pixel identification is performed to identify the noisy pixels. A Pixel with label 2 is considered as noise free and hence, without any restoration process, its original value is retained. Some pixels with "0" or "1" labels may be non-noisy. With a process called similarity inspection, it is possible to accurately detect noisy pixels.

### 3) Partitioning

In order to identify the noisy pixels, similarity among neighboring pixels must be inspected. To this aim for each pixel in a 3×3 window, a 3×3 neighboring window is partitioned and fed to the similarity inspection module.

### 4) Local Similarity Inspection

In this stage, using labels 0, 1 and 2; similarities between neighboring pixels in a 3×3 window are obtained. Pixels with 0 and 1 labels are potentially noisy. For a given pixel which has a label of 0 or 1 its similarity in 3×3 window is computed. If the number of pixels with different labels from the central pixel is greater than a threshold ($T1$), the central pixel is non-similar to its neighbors and hence it is considered as being noisy. The result of similarity inspection is formation of mask which is shown in Fig. 2(d). In other words, if pixels with intensity values of 0 or 255 do not have the same values as their neighbors, they must be noisy pixels. The labeling procedure and the similarity inspection are performed in such a way that edges are preserved.

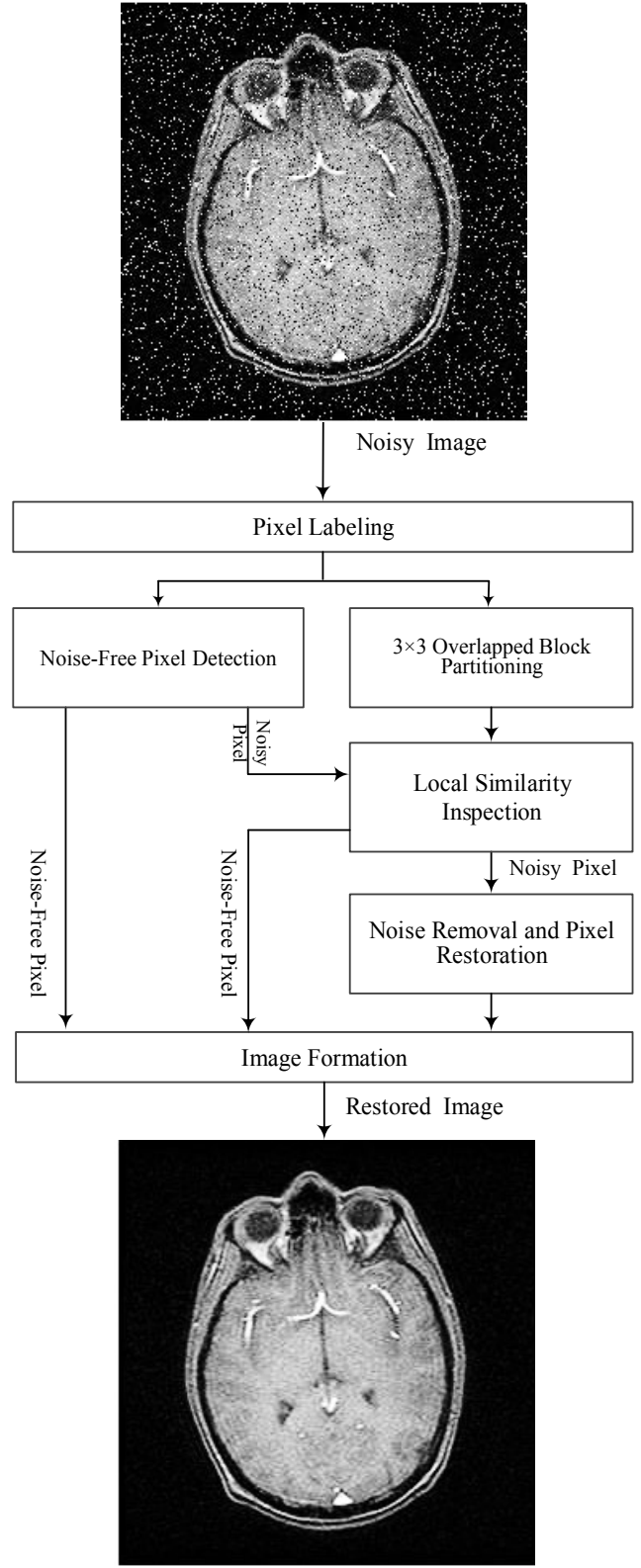

**Fig 1**. Flowchart of the proposed method.

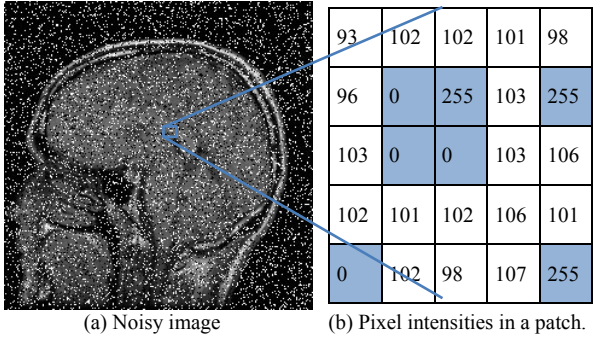

(a) Noisy image  
(b) Pixel intensities in a patch.

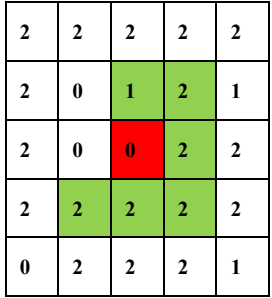
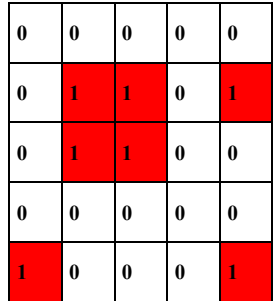

(c) Labeling each pixel with one of the three labels.  
(d) Mask for noisy pixels. 0 for non-noisy and 1 for noisy pixels.

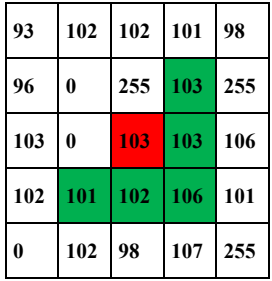
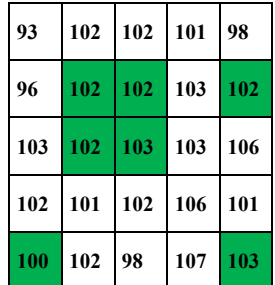

(e) Pixel denoising (Median filter used on non-noisy pixels surrounding the noisy pixel.)  
(f) Results of the denoising process applied to all of window elements. Restored pixels shown in green.

**Fig 2.** Noise detection process on an edge region.

### 5) Noise Removal and Pixel Restoration

In this stage, the noisy pixels, which were identified in the previous stage, are replaced. The noisy pixels imply incorrect information and must be removed from the upcoming decision making of the restoration stage.

This stage of the algorithm is dedicated to replacing the noisy pixels with proper values. Hence, median operator is applied only to non-noisy neighboring pixels. Figure 2(e) shows an example of non-noisy pixels that are surrounding the central noisy pixel. The value found by the median operator is assigned to the central pixel. In this way, original similarity that existed between neighboring pixels is restored. Figure 2(f) shows the result of applying the restoration process to all elements of the window.

### 6) Image Formation

Detected noise-free pixels and restored pixels are placed back to form the noise-free image.

## B. Hardware Structure

The proposed noise removal algorithm is designed such that it is suitable for hardware implementation. Figure 3 shows the main hardware blocks of the proposed method. Different parts and modules of the hardware structure of the proposed algorithm are explained in the followings:

### 1) Pixel labeling and Noise-Free Pixel Detection

In Fig. 4 pixel labeler module is illustrated. A comparator and a multiplexer structure are used for labeling of pixels with labels of 0, 1 and 2. For 255, 0 and noise-free pixels the label 1, 0 and 2 is assigned respectively. As it is shown in Fig. 4, pixels are considered to have 8-bit representations. Hence, comparison with 255 and 0 is possible with an 8-input AND gate and an 8-input NOR gate respectively. Noise-free pixels are labeled as "2" and in the reconstruction stage their original values are retained as the final restored values. Hence, pixels with label "2" are directly transferred to the formation module.

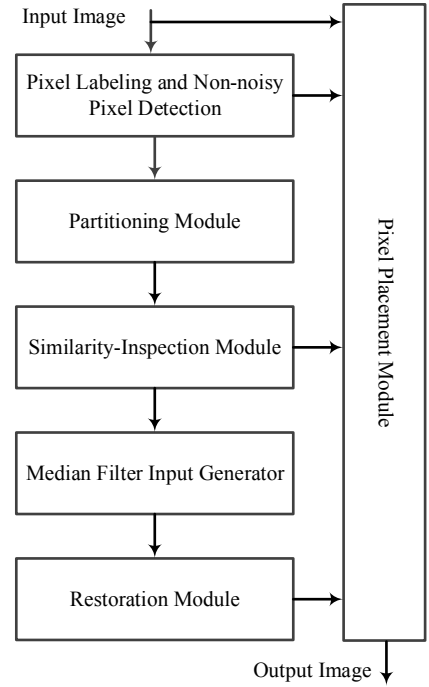

**Fig. 3.** Basic modules of the proposed method.

### 2) Block Partitioning Module

In order to determine if a pixel is noisy, the similarity among pixels, in the vicinity of the pixel, must be considered. This is done by looking at nine $3 \times 3$ blocks around the pixel under the consideration. One of these blocks has this pixel at its center and the other eight blocks have the eight neighboring pixels at their centers. The block partitioning module feeds appropriate pixels to the similarity-inspection block.

*3) Similarity Inspection Module*

In this step the similarity between a pixel and its neighbors is analyzed. For a given pixel with a label, if the number of neighboring pixels with different label is greater than a threshold ($T1$), the pixel is identified as a noisy pixel. The amount of similarity can be measured by a comparator, or a majority circuit, in which a threshold value ($T1$) is used to determine the number of similar neighbors. The hardware structure of the similarity inspection module, consisting of nine similarity inspection units, is shown in Fig. 5. Appropriate pixel labels are transferred to this module and the label is compared with the label of the center pixel. Similarity between the label of the center pixel and the label of a neighboring pixel is computed with a comparator unit (CMP). The number of similar pixels is counted by an adder tree which could add up 9 bits together. Then the result of the adder is compared with a threshold ($T1$) to determine the similarity between a pixel and its neighboring pixels.

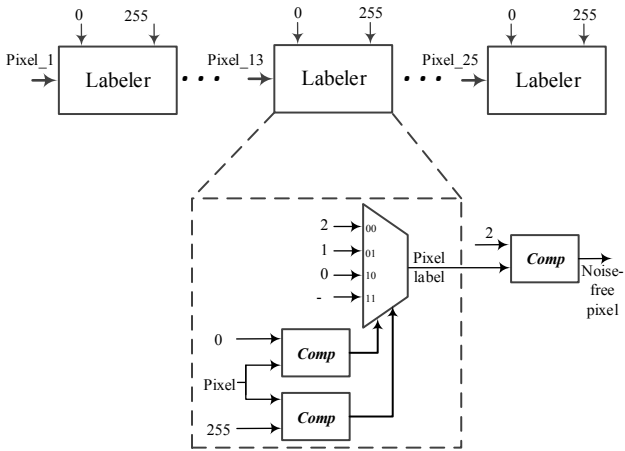

**Fig. 4**. Pixel labeling and noise-free pixel detection.

*4) Median Filter Input Generator*

A noisy pixel should not be involved in the process of restoration of another noisy pixel. This means that we use non-noisy pixels to restore the value of a noisy pixel. Hence, based on results from the similarity inspection module, noisy pixels must be removed from the decision making process of the restoration step.

On the other hand, performing median filter on a variable number of pixel values would increase the complexity. Hence, we place 0 or 255 for the noisy pixels and we know that these pixels will not appear as the median value. The structure shown in Fig. 6 sends out 9 numbers, corresponding to central pixel as well as its 8 neighboring pixels, to a median filter. We call the structure of Fig. 6 as Median Filter Input Generator (MFIG). If a neighboring pixel is non-noisy then its original intensity is sent out by the MFIG. For noisy neighboring pixels, the MFIG sends out 0 or 255. Based on the results from the similarity-inspection module, either the original neighboring pixel's value is sent out or one of the two values of 0 or 255 is output. For noisy pixels, MFIG alternates between 0 and 255 values. The first noisy pixel that MFIG encounters is assigned 0 if the trigger input is 0, otherwise it is assigned 255. Hence, one of the non-noisy pixels is chosen as the median value.

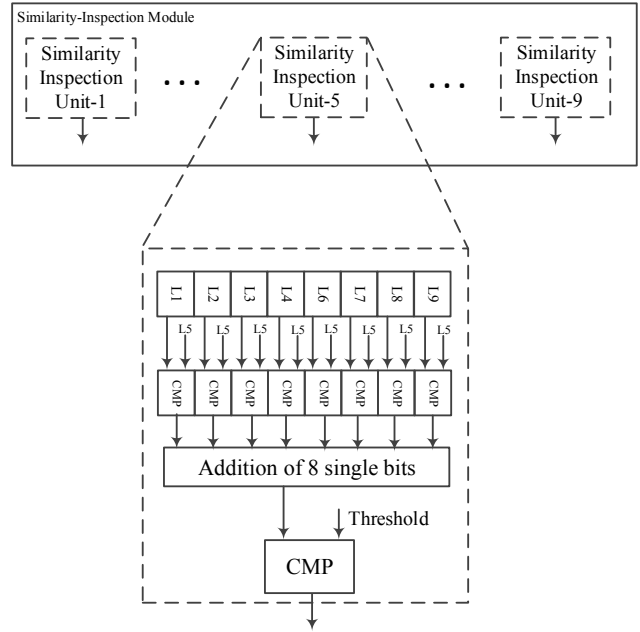

**Fig. 5.** Hardware structure of the similarity-inspection module.

*5) Restoration Module*

If the number of neighboring noisy pixels is odd then MFIG produces one more 0 output when its trigger is 0. On the other hand, if the trigger value of the MFIG is set to 1, it produces one more 255 output as compared to the number of 0 outputs. The list of non-noisy neighboring pixels would shift one position depending on the chosen trigger value. Hence, in Fig. 7 two MFIG units are used to produce both possible lists of inputs for the median filters. The average of the two median values is computed and a rounded value is output in Fig. 7.

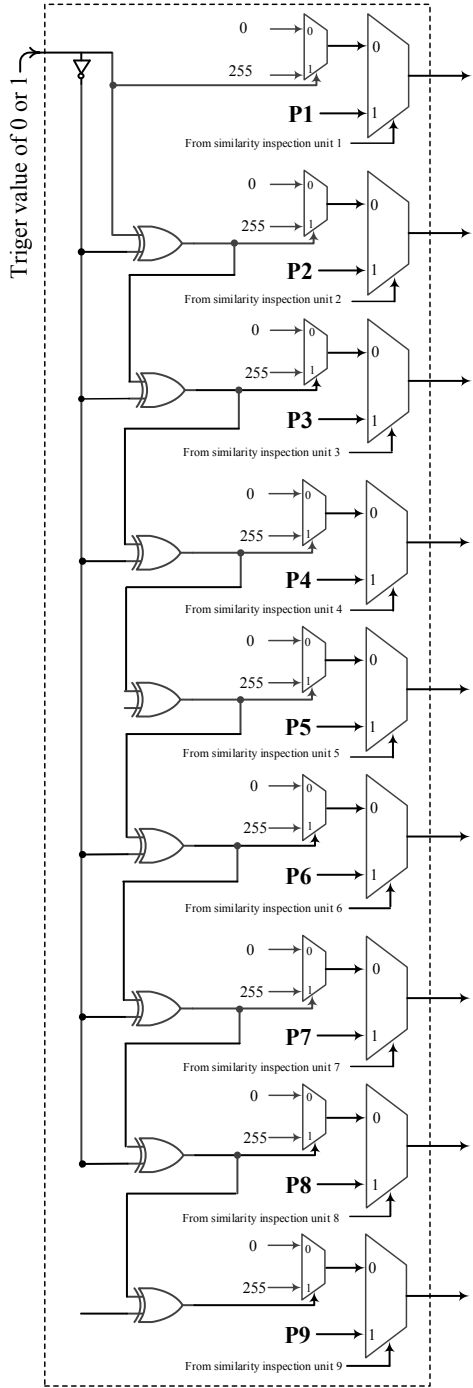

**Fig. 6.** Median filter input generator (MFIG).

*6) Pixel Placement Module*

At the final stage of the proposed hardware pixels are placed in the de-noised image. This is performed by the pixel placement module as shown in Fig. 8. A pixel may have been detected as non-noisy and its original value is placed in the image. Also, for a noisy pixel its restored value is chosen and placed in the de-noised image.

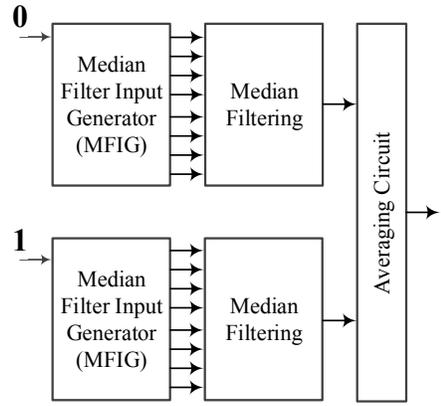

**Fig. 7.** Restoration module

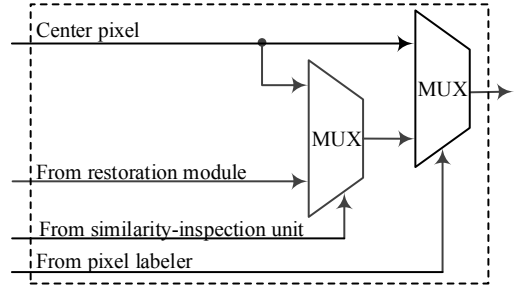

**Fig. 8.** Hardware structure of pixel-placement module

### III. EXPERIMENTAL RESULTS

To verifying the accuracy and show the low complexity of our proposed method, simulations are performed in two stages as follows:

#### A. Software Simulation

Experiments are performed and verified in MATLAB. Firstly for visual quality verification, standard 8-bit gray-scale MR images are used with the size of $256 \times 256$ [22]. Noise density of 20 % is to MR images. To illustrate visual qualities of our method as compared to different denoising techniques, three MR images are selected and results are illustrated in Fig 9, Fig. 10, and Fig. 11. Figures 9, 10, and 11 show that the proposed method is able to identify noisy pixels and it is capable of preserving the original pixel values.

Overall 124 standard 8-bit gray-scale MR images are used with the size of $256 \times 256$ [22]. Impulse noise densities (salt and pepper) from 5% to 20% are injected uniformly. Peak signal to noise ratio (PSNR) is used to assess the quality of the restored images. As indicated in Table 1, the proposed method has better results than comparable methods for all noise densities.

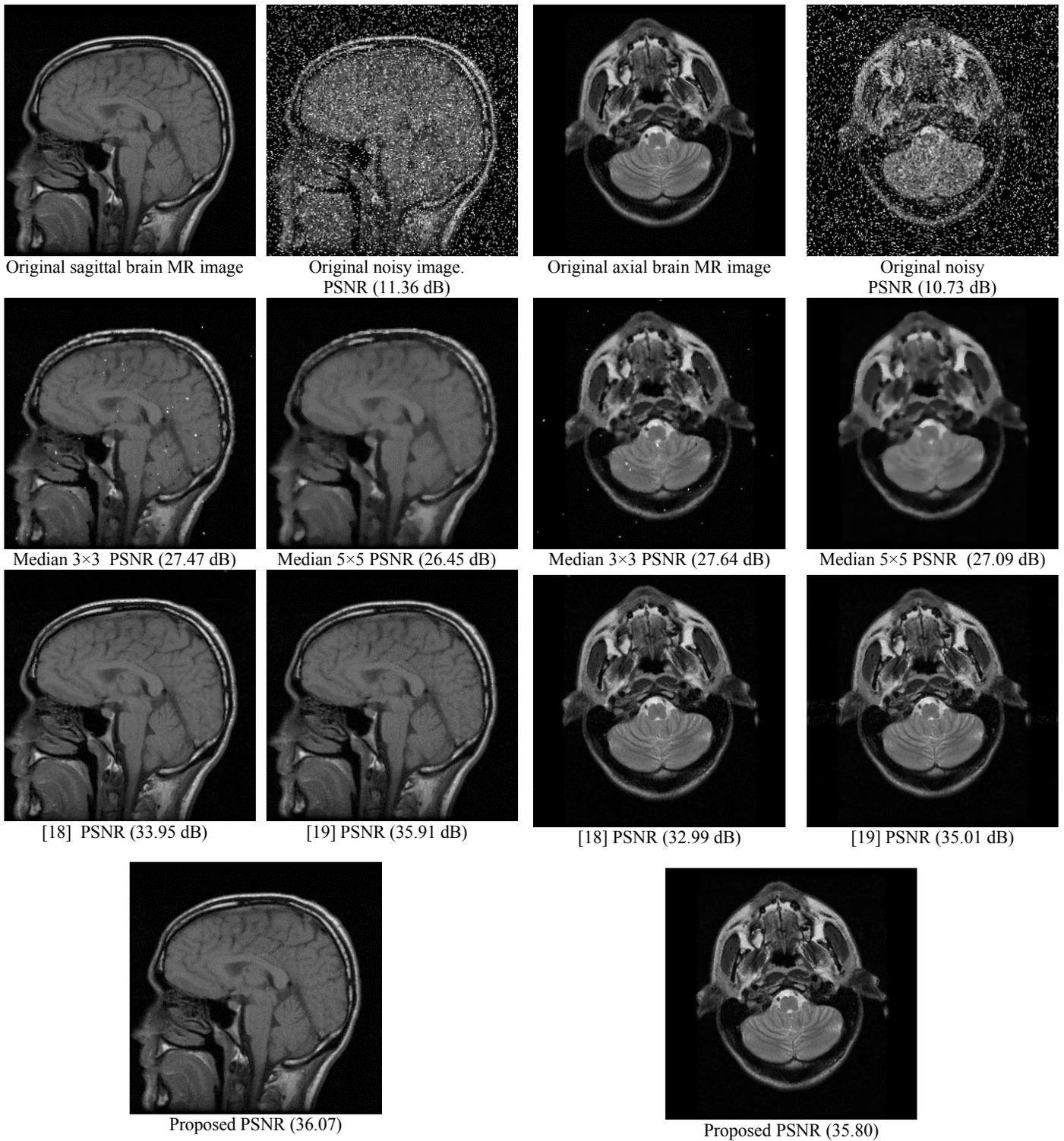

**Fig. 9.** Visual and objective quality measures from our proposed method as compared to median filtering and methods of [18] and [19] for a sagittal brain MR image.

**Fig. 10**. Visual and objective quality measures from our proposed method as compared to median filtering and methods of [18] and [19] for an axial brain MR image.

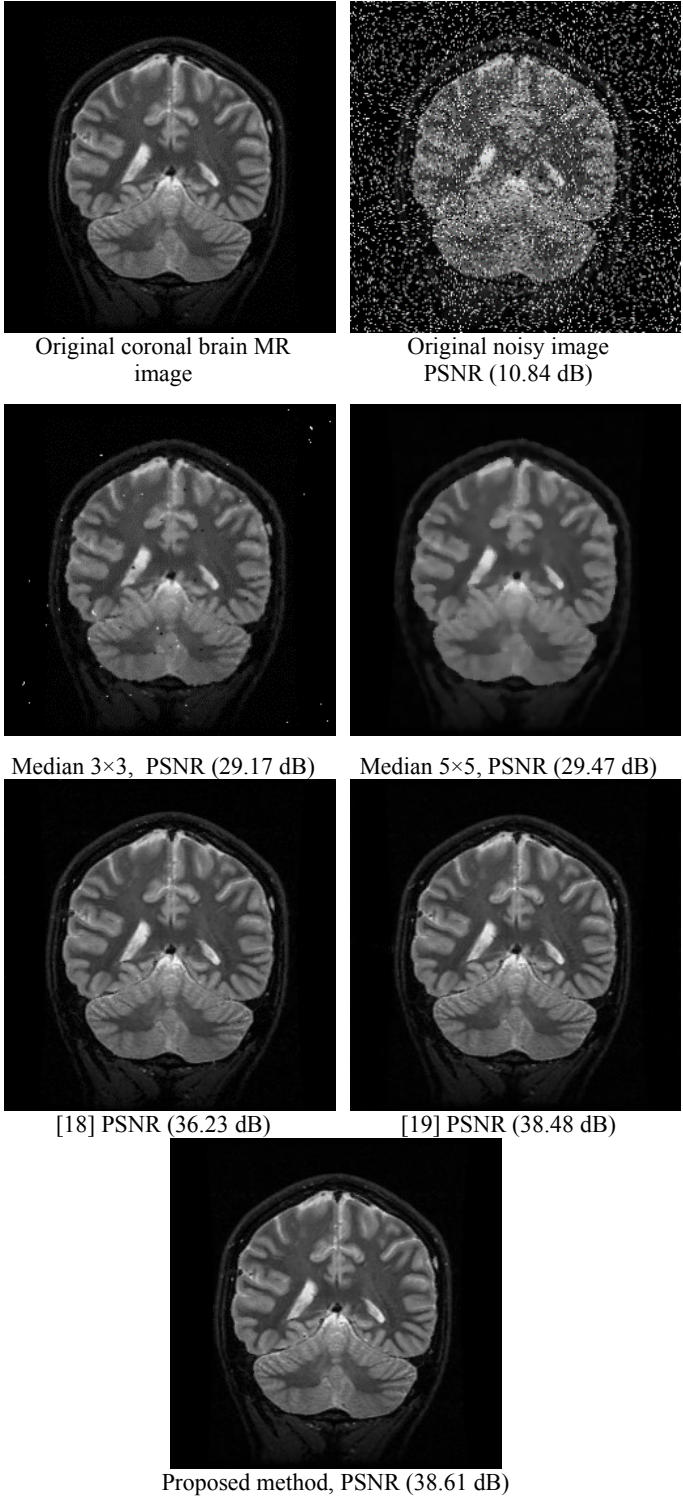

Original coronal brain MR image

Original noisy image PSNR (10.84 dB)

Median 3×3, PSNR (29.17 dB)

Median 5×5, PSNR (29.47 dB)

[18] PSNR (36.23 dB)

[19] PSNR (38.48 dB)

Proposed method, PSNR (38.61 dB)

**Fig. 11.** Visual and objective quality (PSNR) measures from our proposed method as compared to median filtering and methods of [18] and [19] for a coronal brain MR image.

In [19] a comparison between 255 and 0 are used for noise density determination. Naturally pixels with 0 and 255 values exist in the medical images. Hence, only checking to see if a pixel has 0 or 255 for noise detection may lead to a wrong decision. On the other hand, while 0 and 255 could be useful in detection of salt and pepper noisy pixels, authors of [18] do not use these two essential values. In our proposed method, we used both aspects. We know that a noisy pixel is either 0 or 255 but not all 0s and 255 pixels are noisy. Hence, we also inspect the similarity between neighboring pixels. This similarity inspection process causes better de-noising results as compared to other methods. Finally to verifying scalability of the proposed method noise densities from 5% to 25% are added to images. Simulation results, in terms of PSNR values, are illustrated in Fig. 12. Figure 12 shows that the proposed method is capable of de-noising at different noise densities as compared to other comparable methods.

**Table 1**. Comparative results in terms of PSNR (dB) for different noise densities.

| Method | 5% | 10% | 15% | 20% |
|---|---|---|---|---|
| Median(3×3) | 34.29 | 33.23 | 31.29 | 28.52 |
| Median(5×5) | 30.17 | 29.99 | 29.78 | 29.57 |
| [18] | 38.22 | 36.42 | 35.02 | 33.62 |
| [19] | 42.29 | 40.23 | 38.78 | 37.62 |
| Proposed | **43.80** | **41.37** | **39.59** | **38.07** |

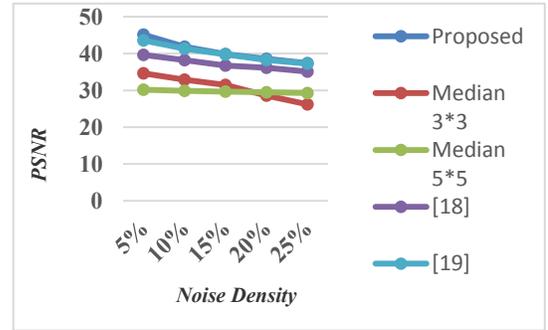

**Fig. 12** Scalability of different methods.

### B. *Complexity Analysis*

For complexity analysis of the proposed method, an FPGA implementation is performed. Proposed architecture is described in VHDL and is implemented on a XILINX Spartan3 family device. The selected target device is xc3sd1800a and hardware implementation is dedicated to MR images with the size of 256×256. An input image is read from the internal RAM and the de-noised image is written back into the same RAM. Implementation results for the FPGA design are summarized and compared in Table 2. In the proposed algorithm all stages, including detection and restoration, are designed to have low hardware complexity. Simple hardware structure is applied for the implementation of the median filter which uses a set of comparators. Low resource utilization, as reported in Table 2, verifies that the proposed denoising algorithm is suitable for hardware implementation in the form of an FPGA device.

## IV. CONCLUSION

In this paper a low complexity noise removal system was proposed for medical images. This method was shown to be suitable for hardware implementation in medical image capturing and transmission devices. The proposed method consisted of two stages of detection and restoration. The goal was to separately improve the accuracy in each of the two stages with respect to hardware complexity. High accuracy of noisy-pixel detection in the first stage, and their removal in the next stage, led to better restoration of noisy images. Simulation results using MATLAB, performed on MR images, showed that the proposed approach removes salt and pepper noise with high accuracy. For each stage of the proposed method an efficient hardware structure is proposed. Low hardware resource utilization of the proposed method shows its suitability to be an integral part of any medical imaging systems.

Table. 2. Hardware utilization comparison.

| Method | Maximum frequency | Area | | | FPGA Device |
|---|---|---|---|---|---|
| | | # of 4input LUTs | # of Slice Flip Flops | # of Slices | |
| Proposed | 181 MHz | 480 (2%) | 1280 (2%) | 1016 (6%) | XILINX Spartan3 |
| Method 1 of [18] (RSEPD) | 162.6 MHz | 709 (Logic Cells) | | | Altera STRATIX EPIS25 |
| Method 2 of [18] (SEPD) | 72.3 MHz | 1487 (Logic Cells) | | | Altera STRATIX EPIS25 |